\documentclass{article}

\usepackage[preprint]{neurips_2025}
\usepackage{comment}

\usepackage{algorithm}
\usepackage{algpseudocode}
\usepackage{enumitem}
\setlist[enumerate]{label=(\alph*),leftmargin=1.5em,topsep=0pt,itemsep=0pt}
\usepackage{graphicx} 
\usepackage{booktabs}
\usepackage{graphicx}
\usepackage[table]{xcolor}   % 用于单元格填色

\usepackage[utf8]{inputenc} % allow utf-8 input
\usepackage[T1]{fontenc}    % use 8-bit T1 fonts

\usepackage{url}            % simple URL typesetting
\usepackage{booktabs}       % professional-quality tables
\usepackage{amsfonts}       % blackboard math symbols
\usepackage{amsmath}        % for \text and math environments
\usepackage{nicefrac}       % compact symbols for 1/2, etc.
\usepackage{microtype}      % microtypography
\usepackage{xcolor}         % colors
\usepackage{fancyhdr}
\usepackage{hyperref}       % hyperlinks
\usepackage{multirow}

\makeatletter
\renewcommand{\thefootnote}{\fnsymbol{footnote}}
\makeatother
\makeatletter
\renewcommand{\@noticestring}{}   % 不再显示任何文字
\makeatother

\title{FastInit: Fast Noise Initialization for Temporally Consistent Video Generation 
}

\author{%
  Chengyu~Bai\textsuperscript{1,2*§}\quad
  Yuming~Li\textsuperscript{1,2*§}\quad
  Zhongyu~Zhao\textsuperscript{1}\quad
  Jintao~Chen\textsuperscript{1}\\
  \textbf{Peidong~Jia}\textsuperscript{1}\quad
  \textbf{Qi~She}\textsuperscript{2$\dagger$}\quad
  \textbf{Ming~Lu}\textsuperscript{1$\dagger$}\quad
  \textbf{Shanghang~Zhang}\textsuperscript{1$\ddagger$}\\[0.4em]
  \textsuperscript{1}Peking~University\quad
  \textsuperscript{2}ByteDance\\[0.6em]
  {\footnotesize
    $^{*}$Equal contribution.\quad
    $^{\dagger}$Project leader.\quad
    $^{\ddagger}$Corresponding author.}%
}

\begin{document}

\maketitle

\begingroup
\renewcommand{\thefootnote}{\fnsymbol{footnote}} % 保持符号编号
\footnotetext[4]{Work done while the author was an intern at ByteDance, supervised by Dr.~Qi~She.}
\endgroup
\newcommand{\bestcell}[1]{\cellcolor{yellow!30}\textbf{#1}} % 浅蓝背景 + 加粗

%%%%%%%%%%%%%%%%%%%%%%%%%%%%%%%%%%%
\begin{abstract}
Video generation has made significant strides with the development of diffusion models; however, achieving high temporal consistency remains a challenging task. Recently, FreeInit identified a training-inference gap and introduced a method to iteratively refine the initial noise during inference. However, iterative refinement significantly increases the computational cost associated with video generation. In this paper, we introduce FastInit, a fast noise initialization method that eliminates the need for iterative refinement. 
%We introduce the concept of golden noise of video, where the initial noise is transformed into a noise distribution enriched with semantic information, thereby improving the coherence and natural dynamics of video frames. 
FastInit learns a Video Noise Prediction Network (VNPNet) that takes random noise and a text prompt as input, generating refined noise in a single forward pass. Therefore, FastInit greatly enhances the efficiency of video generation while achieving high temporal consistency across frames. To train the VNPNet, we create a large-scale dataset consisting of pairs of text prompts, random noise, and refined noise. Extensive experiments with various text-to-video models show that our method consistently improves the quality and temporal consistency of the generated videos. FastInit not only provides a substantial improvement in video generation but also offers a practical solution that can be applied directly during inference. The code and dataset will be released.
\end{abstract}

%%%%%%%%%%%%%%%%%%%%%%%%%%%%%%%%%%%
\section{Introduction}
\begin{figure}
    \centering
    \includegraphics[width=1\linewidth]{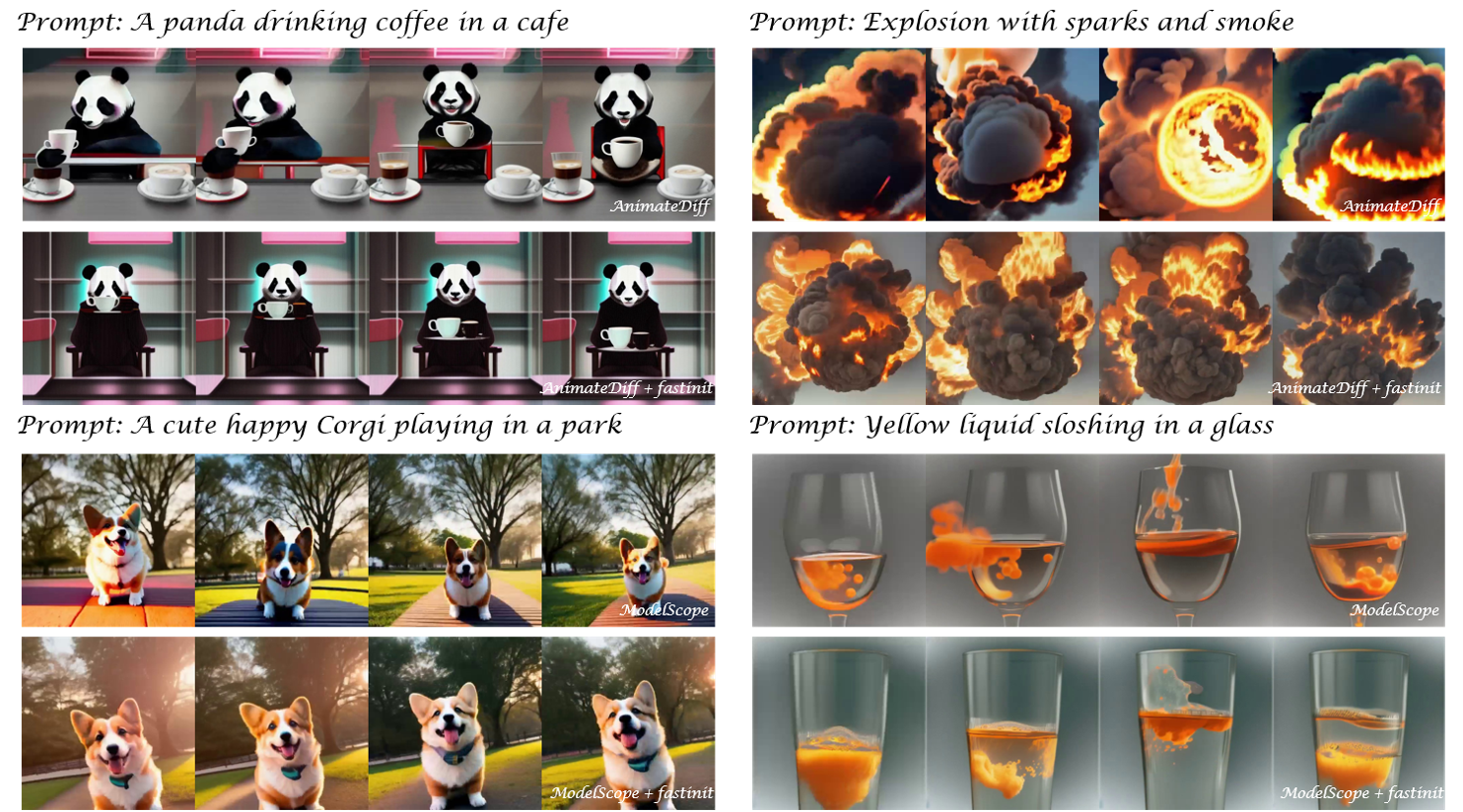}
    \caption{\textbf{Example video generation results} with noises initialized by the proposed FastInit. We propose a method called FastInit, which is a fast noise-initialization scheme that eliminates the need for iterative refinement. Given a random noise and a text prompt, FastInit generates a refined noise in a single forward pass. The resulting videos show significantly improved temporal consistency and visual fidelity, and the method can be integrated into existing video generation models during inference.}
    \label{fig:demo}
\end{figure}
Recent advancements in video generation models, particularly those based on diffusion models, have shown significant progress. Increasing efforts are being made to create visually appealing video clips from textual descriptions through text-to-video (T2V) generation ~\cite{ho2022imagenvideohighdefinition,zhou2023magicvideoefficientvideogeneration,singer2022makeavideotexttovideogenerationtextvideo,wang2023modelscopetexttovideotechnicalreport,blattmann2023stable,girdhar2024factorizing,bao2024vidu,yang2025cogvideoxtexttovideodiffusionmodels,kong2025hunyuanvideosystematicframeworklarge,wang2025wan}. Current video generative models, such as AnimateDiff~\cite{guo2023animatediff}, VideoCrafter ~\cite{chen2024videocrafter2}  typically utilize pre-trained image diffusion models as their foundation, adding temporal layers or motion modeling components to effectively capture temporal dependencies in videos and generate coherent video sequences. Despite their success in producing high-quality videos, these models still encounter challenges, particularly with temporal consistency. Generated videos often display unnatural dynamics, especially in longer sequences.

Existing studies have revealed that the initial noise is crucial in diffusion-based image generation, greatly influencing semantic layout, object composition, and overall image fidelity~\cite{zhou2024golden,ahn2024noise,wu2024freeinit}. Unlike image generation, video generation requires maintaining temporal consistency across frames. Notably, the low-frequency components of the initial noise—often associated with consistent visual elements like backgrounds and primary subjects—tend to remain stable throughout the diffusion process.

% Recent efforts have explored optimizing the initial noise in video diffusion models to improve generation quality. Noise Calibration~\cite{yang2024noise} refines initial noise through iterative denoising guided by a pretrained video diffusion model to enhance visual fidelity while preserving content. IV-Mixed Sampler~\cite{shao2024iv} leverages image diffusion models to generate high-quality frames and integrates them into video diffusion pipelines for improved temporal coherence. FreeInit~\cite{wu2024freeinit} identifies a training-inference initialization gap and proposes an iterative strategy to refine the low-frequency components of the initial noise during inference, leading to better subject appearance and motion consistency across frames.
A growing body of research focuses on optimizing the {\bf initial noise} in video diffusion models to enhance generation quality. Noise Calibration~\cite{yang2024noise} conducts iterative denoising using a pretrained video diffuser, enhancing visual fidelity while maintaining scene content. IV-Mixed Sampler~\cite{shao2024iv} first generates high-quality frames using an image-diffusion backbone, and then integrates the resulting latents into a video pipeline to enhance temporal coherence. FreeInit~\cite{wu2024freeinit} addresses the gap between training and inference by iteratively refining low-frequency noise components during inference. This process enhances the consistency of subject appearance and motion across frames.

Despite recent progress in noise optimization for video generation, existing methods still face several significant limitations. First, they usually depend on iterative refinement, which leads to significant inference overhead, particularly problematic for lengthy video sequences. Second, their designs for specific tasks often limit the ability to generalize across various models and datasets. Third, many approaches require modifications to the backbone model, limiting their practical deployment in real-world applications.

% To address the aforementioned challenges, we propose \textbf{FastInit}, a novel one-shot noise optimization framework that eliminates the need for iterative refinement. At its core, FastInit learns a \textbf{Video Noise Prediction Network (VNPNet)} that takes as input a random noise vector and a text prompt, and outputs a refined noise in a single forward pass. This refined noise embeds semantic priors that guide the generation process toward temporally consistent and visually coherent results. Most entries in a random noise tensor are superfluous for generation; the informative signal resides in the coarse, low-frequency patterns that encode scene structure and the outlines of the main objects.
% To extract this signal, we employ \textbf{Tucker decomposition}—a higher-order analogue of PCA/SVD—that factorises a high-dimensional tensor into mode-specific orthonormal bases and a compact core.
% A lightweight, \emph{factor-aware attention gate} together with \emph{core masking} then functions as a structure-preserving denoiser: it adaptively suppresses useless high-frequency perturbations while retaining the semantically meaningful, low-frequency content essential for stable video generation.

To address the aforementioned challenges, we introduce \textbf{FastInit}, a novel noise-initialisation framework that replaces costly iterative refinement with a single forward pass. At its heart lies the Video Noise Prediction Network (VNPNet), which takes random Gaussian noise and a text prompt to produce a refined noise that enhances temporal consistency. Injecting this refined noise into an off-the-shelf diffusion sampler yields videos that are both temporally consistent and visually coherent. VNPNet is based on the idea that informative signals in noise are found in their coarse, low-frequency components, which capture the global layout of a scene and the silhouettes of primary objects. To isolate the useful signal, we apply a differentiable Tucker decomposition that factorizes the 4-D latent space into mode-specific orthonormal bases and a compact core. On top of this factorisation, a factor-aware attention selectively amplifies or diminishes principal bases, while a learnable core mask reduces irrelevant components. These two mechanisms work together as a denoiser that preserves structure, adaptively reducing high-frequency noise while maintaining the low-frequency content crucial for stable video generation.

To train VNPNet, we create a large-scale dataset of pairs comprising text prompts, random noise, and refined noise. FastInit is model-agnostic and requires no changes to existing diffusion backbones, making it easy to apply during inference, it can be used as a plug-and-play module during inference. Extensive experiments on various text-to-video models show that FastInit enhances visual quality and temporal consistency while significantly decreasing inference time compared to previous noise refinement techniques. As illustrated in Fig.~\ref{fig:demo}, FastInit significantly improves temporal consistency and visual quality in generated frames. In summary, our main contributions can be concluded as follows:

\begin{itemize}
  \item We propose \textbf{FastInit}, a novel and efficient noise initialization framework for video diffusion models, which directly predicts refined initial noise in a single forward pass, eliminating the need for costly iterative refinement.

  \item We design a lightweight yet effective Video Noise Prediction Network (VNPNet) and construct a large-scale Prompt-Noise Dataset (PNData) to enable supervised training of VNPNet.

  \item We perform extensive experiments across various text-to-video backbones, demonstrating that FastInit significantly enhances both visual quality and temporal consistency while decreasing inference time compared to iterative refinement.
\end{itemize}

\begin{figure}
    \centering
    \includegraphics[width=1\linewidth]{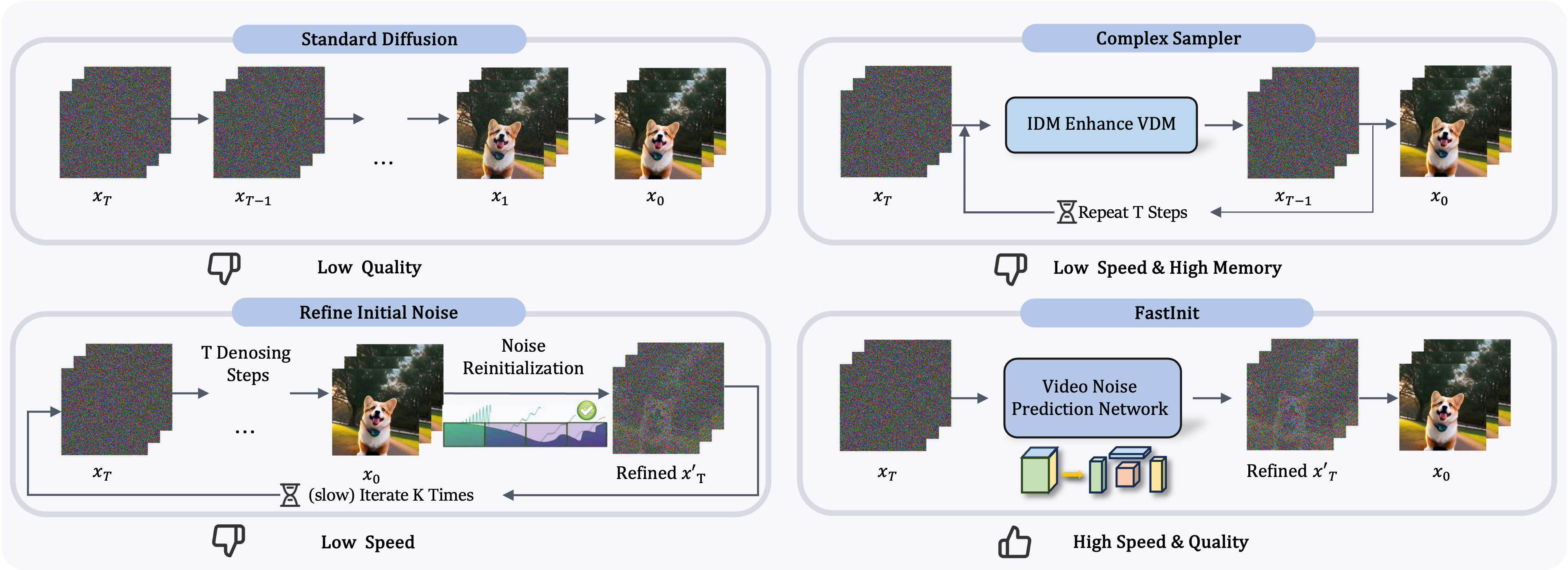}
    \caption{\textbf{The illustration of various pipelines.} Standard diffusion begins with random noise but produces low quality. Refine initial noise improves quality through slow iterative denoising. Complex Sampler uses an image-diffusion backbone for iterative enhancement at denosing step, resulting in low speed and high memory consumption. FastInit employs a prediction network to generate refined noise directly.}
    \label{fig:different_pipelines}
\end{figure}

%%%%%%%%%%%%%%%%%%%%%%%%%%%%%%%%%%%
\section{Related Work}
\label{related work}

{\bf Video diffusion models.} Recent advancements in video generation have been propelled by the introduction of diffusion models into the realm of video. Diffusion models, initially successful in generating high-quality images, have demonstrated significant potential in producing temporally coherent video sequences. Video diffusion models have significantly advanced through the integration of spatio-temporal dependencies, with popular models including AnimateDiff~\cite{guo2023animatediff}, VideoCrafter~\cite{chen2024videocrafter2}, WAN~\cite{wang2025wan}, CogVideo~\cite{hong2022cogvideo}, and OpenSora~\cite{zheng2024open}. 
%These models extend pre-trained text-to-image diffusion frameworks to video, leveraging temporal layers or incorporating motion modeling components to improve coherence and realism across video frames. 
Although video generation has advanced, maintaining temporal consistency in long sequences remains challenging. Our approach aims to improve the quality of video generation by refining the initial noise during inference, thus enhancing temporal consistency without the need to retrain the diffusion model.

{\bf Inference-Time Enhancement.} Researchers have investigated methods to improve the generation quality of diffusion models during the inference stage, inspired by scaling techniques used in large language models~\cite{snell2024scalingllmtesttimecompute}. Among various strategies, rectifying internal components such as rotary position embeddings~\cite{zhao2025riflex}, attention mechanisms~\cite{qiu2023freenoise,lu2024freelong}, and random seeds~\cite{xu2025goodseedmakesgood,li2025seedsequalenhancingcompositional} has proven effective.
Some methods modify the attention layers in transformer blocks to better balance global and local temporal features, allowing short video models to produce longer and more coherent sequences~\cite{lu2024freelong}. Some researchers have incorporated temporal sliding windows into the attention mechanism to enhance local frame consistency~\cite{qiu2023freenoise}. 
There are approaches that identify intrinsic frequency patterns in positional encodings and make minimal adaptations to significantly extend the generation length~\cite{zhao2025riflex}. 
Moreover, changing random seed strategies has demonstrated unexpected advantages in stabilizing the sampling trajectory~\cite{xu2025goodseedmakesgood, li2025seedsequalenhancingcompositional}. 
In addition to these architectural adjustments, improving the initial noise during inference has proven to be an effective, model-agnostic strategy for enhancing generation quality, especially in long-form or compositional contexts.

{\bf Initial Noise Refinement.} Optimizing initial noise is a useful technique for guiding the behavior of diffusion models. One of the earliest methods introduced gradient-based refinement to enhance the alignment between the generated content and conditioning signals~\cite{wallace2023end}. Subsequent methods designed explicit loss functions to improve compositionality and semantic coherence, often utilizing attention mechanisms to update the noise either at initialization or during the early denoising steps~\cite{chefer2023attend,guo2024initno}. Another line of research uses memory-based or retrieval-based strategies to create precomputed noise databases that allow for fast inference-time guidance through similarity matching~\cite{wang2024silent}. These methods enable improved alignment with target prompts without the need for gradient updates or retraining. Recently, noise optimization has been formally understood as a practical implementation of scaling during inference~\cite{ma2025inference}. Within this framework, a unified model has emerged: a verifier assesses sample quality while a search algorithm navigates the noise space to identify an appropriate initialization. This has led to increased interest in latent noise refinement, resulting in a variety of inference-time strategies that effectively balance flexibility and quality.

{\bf Initial Video Noise Refinement.} Refining the initial noise in video diffusion models can be generally classified into two main categories: training-free approaches and training-based approaches. 
%Training-free methods operate without additional learning, making them highly efficient and compatible with a wide range of pretrained models. In video generation, noise priors differ significantly from those in image models due to the need for temporal consistency.
A common strategy involves creating reference content—such as videos or sketches—using language or image-based models, then introducing noise to these anchors to build temporally structured priors~\cite{gu2023reuse,ge2023preserve,guo2024i4vgen,wang2024cono,li2024tuning,li2025training,wu2024freeinit}. Self-generated frames can also act as iterative references to enhance long-range coherence~\cite{meng2022sdeditguidedimagesynthesis,yang2024noise}. Some methods begin with standard Gaussian noise and then apply operations such as repetition~\cite{gu2023reuse,qiu2023freenoise}, shuffling~\cite{wang2024cono,qiu2023freenoise}, or time reversal~\cite{gu2023reuse,wang2024cono} to enforce structural priors.

Training-based methods, on the other hand, focus on learning how to transform random noise into more refined noise. GoldenNoise~\cite{zhou2024golden} and NoiseWorth~\cite{ahn2024noise} train networks to map standard Gaussian noise into optimized distributions. The former focuses on learning a semantically rich “golden” noise via NPNet, while the latter removes the need for classifier-free guidance. These methods achieve high-quality image generation. However, extending training-based noise initialization to video generation models has not been explored, to our knowledge. Our approach aims to bridge this gap by achieving improved temporal consistency without the need to retrain the model.

%%%%%%%%%%%%%%%%%%%%%%%%%%%%%%%%%%%
\section{Method}
\label{sec:method}

% \begin{figure}
%     \centering
%     \includegraphics[width=1\linewidth]{pipe_all.png}
%     \caption{\textbf{The illustration of FastInit.} (a) illustrates the overall architecture of VNPNet, showcasing its two main components: the Tucker-Based Noise Filter (TBNF) and the Global Contextual Residual Module (GCRM). In part (b), the TBNF acts as a structure-preserving denoiser by highlighting the principal spatio-temporal components while reducing high-rank stochastic noise. (c) demonstrates that during inference, NPNet can be seamlessly integrated as a plug-and-play module, effectively boosting the performance of existing models across diverse benchmarks.}
%     \label{fig:pipeline}
% \end{figure}

\begin{figure}[t]
    \centering
    \includegraphics[width=1\linewidth]{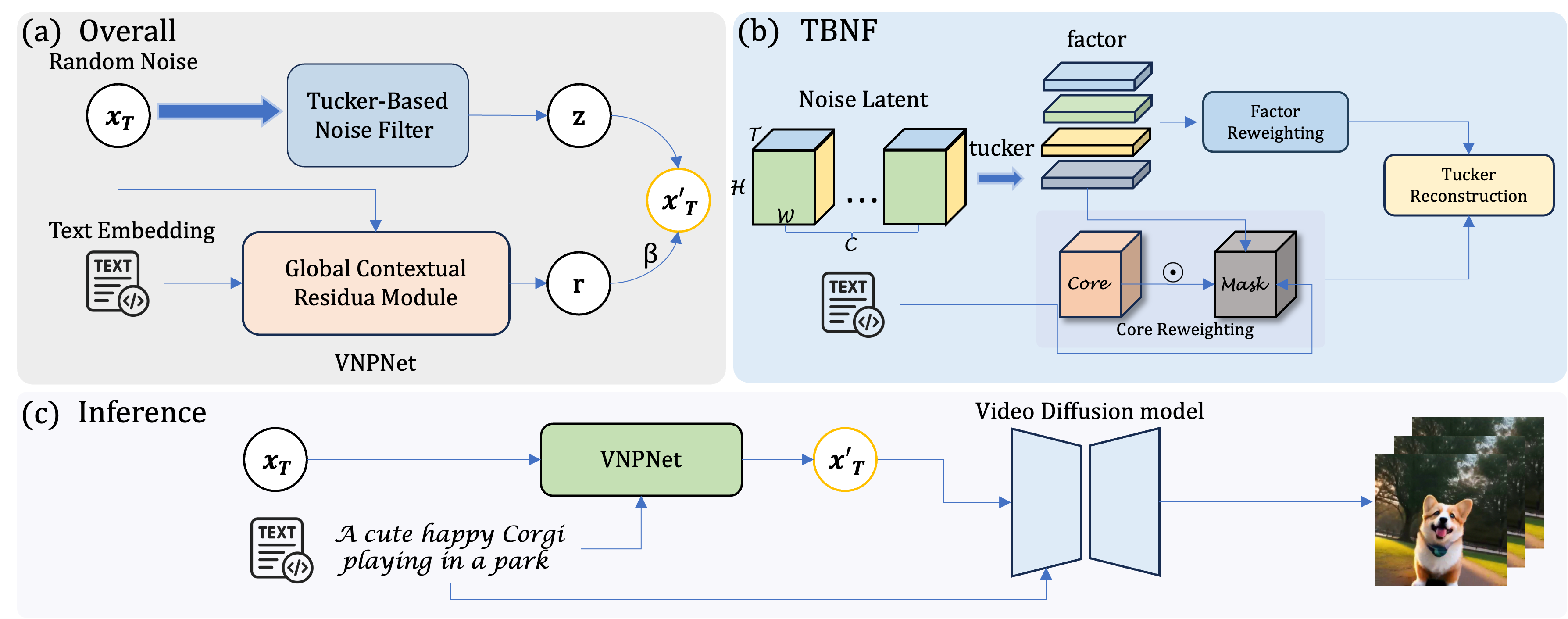}
    \caption{\textbf{The illustration of FastInit.} (a) illustrates the overall architecture of VNPNet, showcasing its two main components: the Tucker-Based Noise Filter (TBNF) and the Global Contextual Residual Module (GCRM). In part (b), the TBNF acts as a structure-preserving denoiser by highlighting the principal spatio-temporal components while reducing high-rank stochastic noise. (c) demonstrates that during inference, NPNet can be seamlessly integrated as a plug-and-play module, effectively boosting the performance of existing models across diverse benchmarks.}
    \label{fig:pipeline}
\end{figure}

While text-to-video diffusion models can produce high-quality frames, they usually start each frame with independent Gaussian noise, which often results in noticeable temporal flicker and motion jitter. To address this limitation, we present FastInit, as illustrated in Fig.~\ref{fig:pipeline}.
(i) We first build a large-scale Prompt Noise Dataset (PNData) (see Section \ref{sec:pnd}), pairing standard i.i.d. Gaussian noise with temporally optimised noise sequences for the same text prompts.
(ii) Leveraging this corpus, we train a Video Noise Prediction Network (VNPNet) (Section \ref{sec:VNPNet}) that maps the random noise directly to a refined latent whose low-frequency structure and inter-frame correlations promote temporally stable and spatially consistent video synthesis.

\subsection{Prompt Noise Dataset (PNData)}
\label{sec:pnd}
FreeInit~\cite{wu2024freeinit} reveals a gap in the initialization between the noise distribution during training and the Gaussian noise utilized during inference. This gap is particularly notable in the low-frequency spectrum, detrimentally affecting temporal coherence and the naturalness of motion.  
To facilitate data-driven research on this phenomenon, we create the Prompt Noise Data (PNData), which is a large-scale pairing of (i) text prompt, (ii) vanilla Gaussian noise, and (iii) noise optimized by FreeInit. As this is not our original contribution, we direct readers to FreeInit~\cite{wu2024freeinit} for additional details.

\subsection{Video Noise Prediction Network (VNPNet)}
\label{sec:VNPNet}
Although GoldenNoise~\cite{zhou2024golden} has been proposed to learn a noise prediction network that maps text prompts and random noise to refine noise for image generation, we find it challenging to extend this strategy to video generation. The main difficulty lies in designing an effective video noise prediction network, as there are significant differences between image generation and video generation. Therefore, we carefully designed the Video Noise Prediction Network (VNPNet) for FastInit.

Let $\mathbf{z}_{\text{rand}}\!\in\!\mathbb{R}^{C\times T\times H\times W}$ denote the initial noise random sampled from
$\mathcal{N}(0,1)$ and
$\mathbf{p}$ is the text embedding of a prompt.
The goal of VNPNet is to learn a mapping
\begin{equation}
\Phi:\;(\mathbf{z}_{\text{rand}},\mathbf{p})\;\longmapsto\;
\mathbf{z}_{\text{refined}},
\end{equation}
where $\mathbf{z}_{\text{refined}}$ is the refined noise, a semantically enriched and temporally coherent latent variable that, when fed into subsequent off-the-shelf diffusion sampler, produces videos that are substantially more stable.
%We realise $\Phi$ with the \textbf{Video Noise Prediction Network} (\textbf{VNPNet}), trained on the Noise Prompt Dataset Collection (\S\ref{sec:npdc}). 
VNPNet $\Phi$ consists of two key components:
\begin{enumerate}
\item \textbf{Tucker-Based Noise Filter (TBNF)}  
      (\S\ref{sec:tucker_pred}).  
      TBNF decomposes random noise via per-sample Tucker factorization, yielding a compact core tensor and four orthonormal mode matrices.
      A set of modality-specific MLP gates re-weights the factor columns, while a learnable core mask sparsifies the high-frequency entries. The reconstruction serves as a structure-preserving denoiser, highlighting key spatio-temporal components while filtering out high-rank stochastic noise.
\item \textbf{Global Contextual Residual Module (GCRM)}  
      (\S\ref{sec:res_pred}).  
      To recover fine details and long-range correlations lost in TBNF, GCRM uses a parameter-efficient UniFormer backbone operating on the text-conditioned latent. Interleaved 3-D convolution and self-attention blocks capture non-local dependencies across frames and modalities, generating a residual volume that globally refines the TBNF output with prompt-aware semantics and coherent motion cues.
\end{enumerate}

Let $\mathbf{e}_{\text{text}} = \textsc{TextEmbed}\!\bigl(\mathbf{z}_{\text{rand}},\mathbf{p}\bigr)$
be the prompt‐conditioned text embedding.
The Tucker branch
$\mathcal{T}(\cdot)$ initially takes only the random noise and exclusively employs $\mathbf{e}_{\text{text}}$ during Core Reweighting.
while the UniFormer residual
$\mathcal{R}(\cdot)$ operates on the text‑enriched hidden state
$\mathbf{z}_{\text{rand}}+\mathbf{e}_{\text{text}}\,$, A learnable parameter $\beta$ is used to control its degree of influence:
\begin{equation}
\label{eq:gold_noise}
\widehat{\mathbf{z}}_{\text{refined}}
  = \underbrace{\mathcal{T}\bigl(\mathbf{z}_{\text{rand}}\bigr)}_{\text{Tucker Filter}}
    \;+\;
    \beta\,
    \underbrace{\mathcal{R}\bigl(\mathbf{z}_{\text{rand}}
                         +\mathbf{e}_{\text{text}}\bigr)}_{\text{Global Contextual}},
\end{equation}
\noindent
where $\beta$ controls the impact of the global contextual term. Adjusting them optimizes the model output $\widehat{\mathbf{z}}_{\text{rand}}$. 

\subsection{Tucker-Based Noise Filter (TBNF)}
\label{sec:tucker_pred}
Let $\mathbf{z}^{(0)}_b\!\in\!\mathbb{R}^{C\times T\times H\times W}$
denote the random noise latent of the $b$‑th sample.
We aim for a structured representation that (i) retains the low-frequency spatio-temporal regularities present in $\mathbf{z}^{(0)}_b$ and (ii) serves as a compact, semantics-aware input for the subsequent denoising process, ultimately producing a temporally coherent latent $\mathbf{z}^{(\ast)}_b$. The raw noise latent is dominated by high-frequency disturbances that provide minimal useful information. By projecting $\mathbf{z}^{(0)}_b$ onto a Tucker manifold, we aim to reduce noisy spikes while maintaining the slowly varying, structurally consistent components that are most informative for stable video generation. For each sample, we compute
\begin{equation}
\label{eq:sample_tucker}
\mathbf{z}^{(0)}_b \;\approx\;
\mathcal{G}_b
\times_1 \mathbf{U}^{(c)}_b
\times_2 \mathbf{U}^{(t)}_b
\times_3 \mathbf{U}^{(h)}_b
\times_4 \mathbf{U}^{(w)}_b,
\end{equation}
\[
\begin{aligned}
\mathbf{U}^{(c)}_b \in \mathbb{R}^{C \times R_c},\
\mathbf{U}^{(t)}_b \in \mathbb{R}^{T \times R_t},\
\mathbf{U}^{(h)}_b \in \mathbb{R}^{H \times R_h},\
\mathbf{U}^{(w)}_b \in \mathbb{R}^{W \times R_w},
\end{aligned}
\qquad
\mathcal{G}_b \in \mathbb{R}^{R_c \times R_t \times R_h \times R_w}.
\]
% where all factor matrices are orthonormal
% ($\mathbf{U}^{(n)\!\top}_b\mathbf{U}^{(n)}_b=\mathbf{I}$).
% Equation~\eqref{eq:sample_tucker}
% separates the latent into a \emph{core–basis} pair that explicitly
% encodes:
% \begin{itemize}\setlength{\itemsep}{2pt}
% \item \emph{Temporal dynamics} via $\mathbf{U}^{(t)}_b$,
%       unveiling coherent motion patterns;
% \item \emph{Spatial geometry} via
%       $\mathbf{U}^{(h)}_b,\mathbf{U}^{(w)}_b$,
%       capturing global layout;
% \item \emph{Semantic abstraction} via $\mathcal{G}_b$,
%       which summarises cross‑modal interactions in a highly compressed
%       form.
% \end{itemize}

The decomposition separates the noise into {\bf various factors and a core}. This formulation clearly encodes the temporal dynamics through $\mathbf{U}^{(t)}_b$, capturing coherent motion patterns; the spatial geometry through $\mathbf{U}^{(h)}_b$ and $\mathbf{U}^{(w)}_b$, which represent the global layout; and the semantic abstraction via $\mathcal{G}_b$, summarizing noise in highly compressed manner.

\paragraph{Factor Reweighting.} For each mode, we apply a multi-layer parametric transformation $\mathcal{N}_{i}$ that maps the statistics of a factor matrix to an attention gate vector using softmax activation:

\begin{equation}
\mathbf{w}_{i}\;=\;
\operatorname{Softmax}\!\bigl(
\mathcal{N}_{i}\bigl[||\mathbf{U}^{(i)}||_{\ell_2}\bigr]\bigr),
\qquad
\mathbf{w}_{i}\in\mathbb{R}^{R_{i}},
\quad i\in{c,t,h,w},
\label{eq:factor_gate}
\end{equation}

% \begin{equation}
% \mathbf{w}_{i}\;=\;
% \operatorname{Softmax}\!\bigl(
%       \operatorname{MLP}_{i}\bigl[\phi\!\bigl(\mathbf{U}^{(i)}\bigr)\bigr]\bigr),
% \qquad 
% \mathbf{w}_{i}\in\mathbb{R}^{R_{i}},
% \quad i\in\{c,t,h,w\},
% \label{eq:factor_gate}
% \end{equation}
where \(\|\cdot\|_{\ell_2}\) indicates element-wise row normalization. Dropout regularization is used in the intermediate feature space to promote generalization.
Each factor matrix is then scaled adaptively using the diagonalized attention weights:

\begin{equation}
\tilde{\mathbf{U}}^{(i)} \;=\;
\mathbf{U}^{(i)} \,\operatorname{diag}\!\bigl(\mathbf{w}_{i}\bigr),
\qquad  i \in c,t,h,w,
\label{eq:factor_scale}
\end{equation}
This adaptive scaling mechanism dynamically emphasizes or suppresses principal components, similar to channel or temporal attention, but operates within the factors of the Tucker manifold.
% \paragraph{Core masking.}
% A learnable mask $\boldsymbol\Gamma
% \!\in\!\mathbb{R}^{1\times R_c\times R_t\times R_h\times R_w}$
% with dropout is applied to the core:
% $\tilde{\mathcal{G}} = \boldsymbol\Gamma \odot \mathcal{G}$.
% Small‐magnitude core elements (often noise) are thus suppressed.
\paragraph{Core Reweighting.} The core tensor $\mathcal{G}_b$ is reweighted by a learnable mask
$\boldsymbol\Gamma \in \mathbb{R}^{1\times R_c\times R_t\times R_h\times R_w}$,
which is constructed through a hierarchical feature integration process that incorporates both factor matrices and text embeddings.
Specifically, we first derive mode-specific features from the factors
$\{{\mathbf{U}}^{(i)}\}_{i\in\{c,t,h,w\}}$, then fuse these features with the prompt embedding \(\mathbf{p}\) via a parameterized transformation network:
\[
\boldsymbol{\Gamma} = \text{Dropout}\left(
    \mathcal{F}\left(
        \left[
            \phi_c\left({\mathbf{U}}^{(c)}\right),\ 
            \ldots,\ 
            \phi_w\left({\mathbf{U}}^{(w)}\right),\ 
            \mathbf{p}
        \right]
    \right)
\right)
\]
The reweighted core is then given by
\[
\tilde{\mathcal{G}} = \boldsymbol\Gamma \odot \mathcal{G}.
\]
This mechanism allows the network to learn a semantic-aware attenuation, minimizing the influence of core entries that are less relevant to the prompt or factors, typically those associated with noisy or weak basis combinations. After performing factor reweighting and core reweighting, the coarse prediction is reconstructed using the following:
\begin{equation}
\widetilde{\mathbf{z}} \;=\;
    \tilde{\mathcal{G}}
    \times_{1} \tilde{\mathbf{U}}^{(c)}
    \times_{2} \tilde{\mathbf{U}}^{(t)}
    \times_{3} \tilde{\mathbf{U}}^{(h)}
    \times_{4} \tilde{\mathbf{U}}^{(w)},
\label{eq:z_reconstruct}
\end{equation}
When projected into the Tucker manifold, the structured content accumulates most of the energy in a few high-magnitude bases. Retaining only these energetically significant bases provides a basic level of noise removal.
%The subsequent factor gates and core mask refine the suppression in a data-adaptive fashion:
%\begin{itemize}\setlength{\itemsep}{2pt}
% \item \textbf{Channel refinement} —  
%       gate $\mathbf{w}_c$ down-scales basis vectors that do not align
%       with prompt semantics, muting spurious channel mixtures;
%\item \textbf{Temporal refinement} —  
%      $\mathbf{w}_t$ emphasises slowly varying patterns and attenuates
%      rapid frame-to-frame oscillations responsible for flicker;
%\item \textbf{Spatial refinement} —  
%      $\mathbf{w}_h$ and $\mathbf{w}_w$ jointly suppress
%      high-frequency spatial bases, eliminating fine-grained speckle
%      noise while preserving global structure.
%\end{itemize}
The cascaded factor reweighting and core reweighting create a differentiable band-pass filter that retains semantically meaningful, low-variance content while adaptively reducing mode-specific noise.

\subsection{Global Contextual Residual Module (GCRM)}
\label{sec:res_pred}
Although the tucker-filter tensor $\widetilde{\mathbf{z}}$ (Sec.~\ref{sec:tucker_pred}) retains semantically meaningful, low-variance content, it also leads to some loss of information. We add a lightweight Global Contextual Refinement Module (GCRM) that utilizes the efficient UniFormer block proposed by~\cite{li2022uniformerunifiedtransformerefficient}.
Within our framework, the GCRM serves two main functions: 1) It retrieves global spatio-temporal context using UniFormer’s mixed 3D convolution and windowed-attention blocks, which aggregate information from distant pixels and frames; 2) It injects prompt semantics by additively fusing a projected text embedding into the video tokens at the initial layer, ensuring that the refinement aligns with the user's prompt~$\mathbf{p}$.

Specifically, the random latent vector \(\mathbf{z}_{\text{rand}}\) is divided into \(P \times P \times P\) tubes and is then flattened into \(N = \frac{THW}{P^3}\) tokens:
\begin{equation}
\mathbf{h}_0 = \phi_{P}\bigl(\mathbf{z}_{\text{rand}}\bigr)\;+\;
           \mathbf{E}_{\text{pos}},
\
\end{equation}
where $\phi_{P}$ is a non-overlapping 3-D patch embed
and $\mathbf{E}_{\text{pos}}$ denotes learnable positional codes.
% ----------------------------------------------------------
% \paragraph{Prompt-aware fusion.}
% Instead of a dedicated cross-attention layer, we broadcast a
% linear projection of the prompt embedding to every token and add it
% to $\mathbf{h}_0$:
% \begin{equation}
% \mathbf{h}_0
% \;\leftarrow\;
% \mathbf{h}_0 \;+\;
% \mathbf{W}_p \mathbf{p},
% \qquad
% \mathbf{W}_p\in\mathbb{R}^{d\times d_{\mathrm{txt}}},
% \end{equation}
% providing a lightweight, prompt-aware bias that guides all subsequent
% UniFormer blocks.
% ----------------------------------------------------------
The token sequence is processed by $L$ stacked UniFormer blocks, with each block consisting of three main components: (i) Dynamic Positional Encoding, implemented through a 3-D depth-wise convolution; (ii) a Multi-Head Relation Aggregator, which utilizes local windows in the initial stages and transitions to global attention in later stages; and (iii) a feed-forward network. Denoting this operator by $\operatorname{Uni}^{(L)}(\cdot)$ we obtain
\begin{equation}
\mathbf{h}_{L} = \operatorname{Uni}^{(L)}(\mathbf{h}_0).
\end{equation}
A transposed patch head $\phi_{P}^{-1}$ folds the tokens back to the latent grid, producing a residual that restores details removed by the
low-rank branch:
\begin{equation}
\mathbf{r} = \phi_{P}^{-1}(\mathbf{h}_{L}), \quad \widehat{\mathbf{z}}_{\text{refined}} = \widetilde{\mathbf{z}} + \beta\mathbf{r}
\end{equation}
% ----------------------------------------------------------
% ChronoMagic-Bench-150
% \begin{table}[t]
% \centering
% \caption{\textbf{Quantitative comparison} on ChronoMagic-Bench (higher is better for UMTScore, MTScore, CHScore; lower is better for UMT-FVD).}
% \label{tab:vbench}
% \begin{tabular}{@{}l l c c c c@{}}
% \toprule
% \textbf{Model} & \textbf{Method} & \textbf{UMT-FVD}\,$\downarrow$ &
% \textbf{UMTScore}\,$\uparrow$ & \textbf{MTScore}\,$\uparrow$ &
% \textbf{CHScore}\,$\uparrow$ \\
% \midrule
% \multirow{Animatediff}
% & Standard DDIM          & 271.36 & 2.63 & 0.43 & 41.98 \\
% & FreeInit               & 254.12 & 2.68 & 0.42 & 48.69 \\
% & IV-mixed Sampler       & 235.73 & \bestcell{2.81} & 0.43 & 44.75 \\
% & \textbf{FastInit (ours)} & \bestcell{233.84} & 2.74 & \bestcell{0.47} & \bestcell{57.85} \\
% \midrule
% \multirow{ModelScope}
% & Standard DDIM          & 254.86 & 2.71 & 0.40 & 51.32 \\
% & FreeInit               & \bestcell{233.72} & 2.83 & 0.44 & 54.87 \\
% & IV-mixed Sampler       & 241.12 & 2.82 & 0.42 & 50.74 \\
% & \textbf{FastInit (ours)} & 238.49 & \bestcell{2.83} & \bestcell{0.46} & \bestcell{59.45} \\
% \bottomrule
% \end{tabular}
% \end{table}

%%%%%%%%%%%%%%%%%%%%%%%%%%%%%%%%%%%
\section{Experiments}
\label{sec:experiments}

\begin{table*}[t]
\centering
\label{tab:chronomagic}
\caption{\textbf{Quantitative comparison} with FreeInit and IV-mixed Sampler on Chronomagic-Bench-1649 based on AnimateDiff and ModelScopeT2V.}
\resizebox{\textwidth}{!}{%
\begin{tabular}{llcccc}
\toprule
\textbf{Model} & \textbf{Method} & \textbf{UMT-FVD} $\downarrow$ & \textbf{UMTScore} $\uparrow$ & \textbf{MTScore} $\uparrow$ & \textbf{CHScore} $\uparrow$ \\
\midrule
\multirow{4}{*}{AnimateDiff} % ① rowspan=4 行, * 表示自动宽度
  & Standard DDIM       & 223.58 & 2.81 & 0.45 & 63.02 \\
  & FreeInit (Iter5)    & 211.34 & 2.82 & 0.41 & 66.35 \\
  & IV-mixed Sampler    & \bestcell{196.80} & \bestcell{2.95} & 0.47 & 56.85 \\
  & \textbf{FastInit (ours)} & 198.23 & 2.87 & \bestcell{0.48} & \bestcell{68.42} \\
\midrule
\multirow{4}{*}{ModelScope}
  & Standard DDIM       & 203.54 & 2.83 & 0.38 & 56.93 \\
  & FreeInit (Iter5)    & 197.89 & 2.91 & 0.38 & 60.32 \\
  & IV-mixed Sampler    & 198.76 & 3.01 & 0.41 & 54.74 \\
  & \textbf{FastInit (ours)} & \bestcell{197.43} & \bestcell{3.06} & \bestcell{0.43} & \bestcell{62.45} \\
\bottomrule
\end{tabular}
}
\label{tab:quanti_chronomagic}
\end{table*}
\subsection{Implementation Details}
\label{sec:impl}
{\bf VNPNet configuration.} The Tucker-Based Noise Filter maintains fixed decomposition ranks $[R_c,R_t,R_h,R_w]=[4,\,8,\,32,\,32]$ throughout all experiments. The Global Contextual Residual Module employs the variant UniFormer, More implementation details can be found in \ref{app:imp}.

{\bf Training protocol.} We use AnimateDiff\cite{guo2023animatediff} and ModelScopeT2V\cite{wang2023modelscopetexttovideotechnicalreport} as the baseline video generation models. The VNPNet is trained for 100 iterations on 8 NVIDIA A100 80GB GPUs using AdamW ($\beta_{1}{=}0.9,\;\beta_{2}{=}0.999,\;\mathrm{wd}{=}0.01$), batch size 144, and cosine LR decay from $2{\times}10^{-4}$ to $1{\times}10^{-6}$. The training process requires approximately one day to complete.

{\bf Datasets and Benchmarks.} VNPNet is trained on our PNData(Section ~\ref{sec:pnd}).
We present performance results on two recent evaluation suites: 1) VBench\cite{zheng2025vbench20advancingvideogeneration} – a comprehensive benchmark of user studies and automatic metrics for assessing text-to-video quality; 2) ChronoMagic-Bench\cite{yuan2024chronomagicbenchbenchmarkmetamorphicevaluation}, which specifically focuses on temporal coherence, measuring Temporal LPIPS, FVD-T, and flicker rate. More benchmarks details can be found in \ref{app:benchmark}.

\subsection{PNData Statistics}
\label{sec:pnd_stat}
PNData‑v1 is created on the MSR‑VTT 9k \cite{xu2016msr} datasets set and a random subset of WebVid‑2M~\cite{webvid}, using AnimateDiff and ModelScopeT2V($512^2$, 16 frames) as the backbone. We constructed approximately 150,000 noise pairs for each model.
All tensors are stored in FP32 and Zstandard‑2 compression. The Microsoft Research Video–to–Text (MSR‑VTT) dataset~\cite{xu2016msr} is one of the most widely used open‑domain video–caption benchmarks. It contains 10,000 web video clips from 20 high‑level categories, each clip being transcribed by Amazon Mechanical Turk workers with 20 independent English sentences. The resulting corpus spans roughly 29,000 unique words.

% \begin{table}[h]
%   \centering
%   \caption{\textbf{PNData‑v1 statistics} after \textbf{5} FreeInit iterations.
%   Higher $\uparrow$ / lower $\downarrow$ indicates better quality.}
%   \label{tab:npdc_stats}
%   \begin{tabular}{ccccccc}
%     \toprule
%     \ & Model &Prompts_Num & Iter &
%     Avg.\ $\Delta$DINO$\uparrow$ &
%     Avg.\ $\Delta$FVD$\downarrow$ \\
%     \midrule
%      AnimateDiff &150\,580 & 5 & +4.6 & $-263$ \\
%      ModelScope  &146\,380 & 5 & +2.9 & $-64$
%     \bottomrule
%   \end{tabular}
% \end

\begin{table}[h]
  \centering
  \caption{\textbf{PNData‑v1 statistics} after \textbf{5} FreeInit iterations. Higher $\uparrow$ / lower $\downarrow$ indicates better quality.}
  \label{tab:npdc_stats}
  \begin{tabular}{lccccc} % 首列设为左对齐（l），其余列居中（c）
    \toprule
    Model & Prompts\_Num & Iter & Avg.\ $\Delta$DINO$\uparrow$ & Avg.\ $\Delta$FVD$\downarrow$ \\
    \midrule
    AnimateDiff & 150\,580 & 5 & +4.6 & $-263$ \\
    ModelScope  & 146\,380 & 5 & +2.9 & $-74$ \\
    \bottomrule
  \end{tabular}
\end{table}

\subsection{Experimental Results}

% \textbf{Main Quantitative Experiments}. We evaluate the proposed {VNPNet} on two representative text-to-video (T2V) generation backbones: \texttt{AnimateDiff} and \texttt{ModelScopeT2V}. As shown in Table~\ref{tab:vbench}, we compare our method \texttt{FastInit} with two recent noise optimization baselines: \texttt{FreeInit} and \texttt{IV-Mixed Sampler}. The evaluation is conducted on Chronomagic-Bench at a resolution of \texttt{512×512} with \texttt{16 frames} per clip.

% We report four widely adopted metrics: {UMT-FVD} (lower is better), {UMTScore}, {MTScore}, and {CHScore} (higher is better), which jointly reflect perceptual quality, motion-text alignment, and temporal consistency.

% Across both backbones, our \texttt{FastInit} consistently achieves the best or second-best performance on all metrics, significantly outperforming standard DDIM sampling and demonstrating competitive or superior results compared to \texttt{FreeInit} and \texttt{IV-Mixed}. Notably, our method achieves the best {CHScore} on both models, indicating enhanced coherence between visual and textual content.
\textbf{Quantitative Results}. We assess the FastInit method using two popular text-to-video (T2V) generation frameworks: AnimateDiff and ModelScopeT2V. As shown in Table~\ref{tab:quanti_chronomagic}, we compare our method against two recent noise optimization baselines, FreeInit and IV-Mixed Sampler, conducting experiments on Chronomagic-Bench at a resolution of 512×512 with 16 frames per clip.

We report four widely adopted metrics— UMT-FVD, UMTScore, MTScore, and CHScore —that jointly assess perceptual quality, text-motion alignment, and temporal coherence. Across both backbones, FastInit consistently achieves either the best or second-best scores across all metrics. Notably, it obtains the highest CHScore on both AnimateDiff and ModelScopeT2V, highlighting its effectiveness in enhancing visual-textual alignment. Moreover, it delivers competitive results on UMT-FVD and UMTScore, while achieving a strong overall balance across all evaluation dimensions—outperforming the standard DDIM baseline by a substantial margin.

\begin{table*}[ht]
\centering
\caption{\textbf{Quantitative comparison} with FreeInit and IV-mixed Sampler on VBench across consistency and temporal metrics}
\resizebox{\textwidth}{!}{%
\begin{tabular}{lccccc}
\toprule
\textbf{Method} & \textbf{Subject Consistency} & \textbf{Overall Consistency} & \textbf{Motion Smoothness} & \textbf{Temporal Flickering} & \textbf{Temporal Style} \\
\midrule
\multicolumn{6}{c}{\textbf{AnimateDiff}} \\
\midrule
Standard DDIM & 95.82 & 27.27 & 97.82 & 98.59 & 25.07 \\
Freeinit (Iter5) & \bestcell{96.45} & 27.17 & 98.04 & \bestcell{98.74} & 24.72 \\
IV-mixed Sampler & 92.78 & 26.89 & 96.46 & 97.12 & 24.81 \\
\textbf{FastInit (ours)} & 96.36 & \bestcell{27.47} & \bestcell{98.23} & 98.73 & \bestcell{25.86} \\
\midrule
\multicolumn{6}{c}{\textbf{ModelScope}} \\
\midrule
Standard DDIM & 89.63 & 25.37 & 95.82 & 98.13 & 25.53 \\
Freeinit (Iter5) & 92.84 & 26.03 & \bestcell{96.67} & 98.50 & 26.01 \\
IV-mixed Sampler & 88.19 & 25.73 & 93.73 & 97.41 & 25.80 \\
\textbf{FastInit (ours)} & \bestcell{93.49} & \bestcell{26.12} & 96.43 & \bestcell{98.67} & \bestcell{26.13} \\
\bottomrule
\end{tabular}
}
\label{tab:quanti_vbench}
\end{table*}

Beyond Chronomagic-Bench, we further assess temporal fidelity using the VBench benchmark, as summarized in Table~\ref{tab:quanti_vbench}. This evaluation includes five criteria: Subject Consistency, Overall Consistency, Motion Smoothness, Temporal Flickering, and Temporal Style, which together measure temporal stability, identity preservation, and stylistic coherence.

FastInit continues to demonstrate consistent improvements on both backbones. On AnimateDiff, it achieves the best performance in Overall Consistency, Motion Smoothness, and Temporal Style, showcasing its ability to generate fluid and coherent motion. On ModelScopeT2V, it ranks first in four out of five metrics, notably excelling in Subject Consistency and Temporal Flickering, reflecting superior identity preservation and temporal smoothness. These results affirm that FastInit not only improves perceptual and semantic quality but also substantially enhances the temporal realism of generated videos.

\textbf{Why FastInit Outperforms FreeInit.}
While our model is trained using data produced by FreeInit, we observe that our FastInit achieves higher scores than FreeInit itself across multiple metrics (Table~\ref{tab:freeinit_vs_fastinit}). We believe the reason is that FreeInit generates refined noise through iterative low-frequency adjustments, which fail to incorporate holistic temporal modeling. In contrast, our VNPNet is designed to generalize across various prompt-noise pairs and integrates additional semantic and contextual cues using TBNF and GCRM. As a result, FastInit leverages the FreeInit signal while enhancing it, achieving improved consistency and perceptual quality through learned spatio-temporal priors as shown in Table~\ref{tab:freeinit_vs_fastinit}.

% \begin{table}[h]
% \centering
% \caption{\textbf{Quantitative comparison} of key temporal‑quality metrics with FreeInit;}
% \begin{tabular}{lcccc}
% \toprule
% Method & Subject Consistency & Motion Smoothness & Overall Consistency & Temporal Style \\
% \midrule
% Freeinit Iter1 & 95.82 & 97.82 & 27.27 & 25.07 \\
% Freeinit Iter2 & 95.82 & 98.15 & 27.36 & 25.27 \\
% Freeinit Iter3 & \bestcell{97.01} & 98.22 & 26.97 & 25.05 \\
% Freeinit Iter4 & 96.79 & 98.00 & 27.29 & 24.83 \\
% Freeinit Iter5 & 96.45 & 98.04 & 27.17 & 24.72 \\
% \textbf{FastInit(ours)} & 96.36 & \bestcell{98.23} & \bestcell{27.47} & \bestcell{25.86} \\
% \bottomrule
% \end{tabular}
% \label{tab:key_metrics_blue}
% \end{table}

\begin{table*}[ht]
\centering
\caption{Comparison of FreeInit (Iter1–Iter5) and FastInit across consistency and temporal metrics}
\resizebox{\textwidth}{!}{%
\begin{tabular}{lccccc}
\toprule
\textbf{Method} & \textbf{Subject Consistency} & \textbf{Overall Consistency} & \textbf{Motion Smoothness} & \textbf{Temporal Flickering} & \textbf{Temporal Style} \\
\midrule
FreeInit Iter1     & 95.82 & 27.27 & 97.82 & 98.59 & 25.07 \\
FreeInit Iter2     & 95.82 & 27.36 & 98.15 & 98.51 & 25.27 \\
FreeInit Iter3     & \bestcell{97.01} & 26.97 & 98.22 & 98.70 & 25.05 \\
FreeInit Iter4     & 96.79 & 27.29 & 98.00 & 98.70 & 24.83 \\
FreeInit Iter5     & 96.45 & 27.17 & 98.04 & \bestcell{98.74} & 24.72 \\
FastInit (ours)    & 96.36 & \bestcell{27.47} & \bestcell{98.23} & 98.73 & \bestcell{25.86} \\
\bottomrule
\end{tabular}
}
\label{tab:freeinit_vs_fastinit}
\end{table*}

\textbf{Ablation Studies.}
To evaluate the effectiveness of our Tucker-Based Noise Filter (TBNF), we perform ablation studies by replacing it with two alternatives: a standard MLP-based predictor without structural decomposition, and a factorization module based on SVD. As shown in Table~\ref{tab:ablation}, the Tucker variant demonstrates the best or comparable performance across all metrics, particularly in CHScore and UMTScore. This highlights the benefits of Tucker decomposition in maintaining structured spatio-temporal priors and improving video-text coherence. In contrast, the SVD baseline produces competitive results, but it slightly underperforms in terms of temporal quality. Meanwhile, the MLP variant shows the largest drop, confirming the importance of structured filtering.

\begin{table}[h]
\centering
\caption{\textbf{Ablation Studies} with Animatediff on Chronomagic-Bench-150.}
\begin{tabular}{lcccc}
\toprule
Method & UMT‑FVD$\;\downarrow$ & UMTScore$\;\uparrow$ & MTScore$\;\uparrow$ & CHScore$\;\uparrow$ \\
\midrule
AnimateDiff & \bestcell{232.45} & 2.63 & 0.45 & 41.98 \\
MLP        & 243.14            & 2.53 & \bestcell{0.47} & 50.88 \\
SVD         & 233.63            & 2.72 & 0.42 & 52.65 \\
Tucker      & 233.84            & \bestcell{2.74} & \bestcell{0.47} & \bestcell{57.85} \\
\bottomrule
\end{tabular}
\label{tab:ablation}
\end{table}

\subsection{Computational Complexity}
\label{sec:com}
To assess the efficiency of our method, we compare the inference time and memory consumption of FastInit with existing methods, utilizing AnimateDiff. All experiments adopt FP32 precision, and the IV-mixed-Sampler uses its official default settings. As shown in Table~\ref{tab:efficiency}, FreeInit has an inference time that is over five times longer due to its iterative optimization process, even though it uses the same number of sampling steps. Similarly, the IV-mixed-Sampler also requires more than four times the time. In contrast, FastInit requires only one forward pass and achieves inference speed similar to AnimateDiff.

While our method employs a lightweight VNPNet, the peak memory usage only slightly increases (approximately 1.5 GiB), remaining within a practical range for modern GPUs. This makes FastInit a more scalable and deployment-friendly solution, especially in latency-sensitive scenarios like real-time video synthesis or on-device generation.

\begin{table}[h]
    \centering
    \caption{\textbf{Inference time and peak memory usage} with AnimateDiff for different methods.}
    \begin{tabular}{lcc}
        \toprule
        \textbf{Method / Configuration} & \textbf{Inference Time (s)} & \textbf{Peak Memory} \\
        \midrule
        AnimateDiff                        & 42.38 & 11\,791\,MiB ($\approx$\,11.52\,GiB) \\
        FreeInit (5\,Iters, 25\,Steps)     & 210.96 & 11\,791\,MiB ($\approx$\,11.52\,GiB) \\
        IV-mixed Sampler (25\,Steps)       & 180. 23& 16\,817\,MiB ($\approx$\,16.42\,GiB) \\
        FastInit(ours) (25\,Steps)         & \bestcell{42.47} & \bestcell{13\,404\,MiB ($\approx$\,13.09\,GiB)} \\
        \bottomrule
    \end{tabular}
\label{tab:efficiency}
\end{table}

\section{Conclusion}
In this paper, we introduce a novel noise initialization method called FastInit for video generation. FastInit utilizes a specially designed Video Noise Prediction Network (VNPNet) to generate refined noise using both the text prompt and random noise in a single forward pass. FastInit is significantly more efficient than iterative methods like FreeInit, while also achieving superior results. To train the VNPNet, we also introduce a large-scale Prompt-Noise Dataset (PNData) that features pairs of text prompts, random noise, and refined noise. We hope our work can inspire future works on noise initialization in video generation models.

% \begin{figure}[t]
%     \centering
%     \includegraphics[width=0.3\linewidth]{radar_pretty.png}
%     \caption{Comparison of Freeinit}
%     \label{fig:radar}
% \end{figure}
% \label{sec:method}

\clearpage
\bibliography{references}{}
\bibliographystyle{plain}% You will need to create a references.bib file to include your references

% \section*{References}

% References follow the acknowledgments in the camera-ready paper. Use unnumbered first-level heading for
% the references. Any choice of citation style is acceptable as long as you are
% consistent. It is permissible to reduce the font size to \verb+small+ (9 point)
% when listing the references.
% Note that the Reference section does not count towards the page limit.
% \medskip

% {
% \small

% [1] Alexander, J.A.\ \& Mozer, M.C.\ (1995) Template-based algorithms for
% connectionist rule extraction. In G.\ Tesauro, D.S.\ Touretzky and T.K.\ Leen
% (eds.), {\it Advances in Neural Information Processing Systems 7},
% pp.\ 609--616. Cambridge, MA: MIT Press.

% [2] Bower, J.M.\ \& Beeman, D.\ (1995) {\it The Book of GENESIS: Exploring
%   Realistic Neural Models with the GEneral NEural SImulation System.}  New York:
% TELOS/Springer--Verlag.

% [3] Hasselmo, M.E., Schnell, E.\ \& Barkai, E.\ (1995) Dynamics of learning and
% recall at excitatory recurrent synapses and cholinergic modulation in rat
% hippocampal region CA3. {\it Journal of Neuroscience} {\bf 15}(7):5249-5262.
% }

%%%%%%%%%%%%%%%%%%%%%%%%%%%%%%%%%%%%%%%%%%%%%%%%%%%%%%%%%%%%
\clearpage
\appendix

\section{Technical Appendices and Supplementary Material}
\subsection{Implementation Details}
\label{app:imp}
The Tucker-Based Noise Filter maintains fixed decomposition ranks $[R_c,R_t,R_h,R_w]=[4,\,8,\,32,\,32]$ throughout all experiments.
With the latent size $C{\times}T{\times}H{\times}W
 = 4{\times}16{\times}64{\times}64$,
the core tensor contains $32\,768$ values, achieving a $7{:}1$ compression ratio relative to the full latent space($262\,144$ values).
This configuration was selected through preliminary grid searches to optimize the trade-off between detail preservation and computational efficiency.

The Global Contextual Residual Module incorporates a modified UniFormer\cite{li2022uniformerunifiedtransformerefficient} architecture with: depth $[5,\,8,\,20,\,7]$, embedding dimensions  $[64,\,128,\,320,\,512]$, fixed head dimension 64, and stochastic depth rate 0.30. 
To ensure dimensional compatibility, the input channel parameter \texttt{in\_chans} is set to 4, matching the bottleneck latent dimension. 
The multi-head relation attention (MHRA) mechanism initially processes local contexts through $5{\times}5{\times}5$ window in the first two stages, then transitions to global $8{\times}8{\times}4$ window operations in subsequent processing stages. 
\subsection{Benchmarks Details}
\label{app:benchmark}
\paragraph{VBench.}VBench\cite{zheng2025vbench20advancingvideogeneration} is a comprehensive benchmark for evaluating text-to-video (T2V) generation models, The latest VBench-2.0 version extends evaluation from Superficial Faithfulness (visual quality) to Intrinsic Faithfulness (logical/physical correctness). VBench evaluates text-to-video models through 16 metrics across two categories: (1) Video Quality (subject/background consistency, motion smoothness, dynamic degree, frame quality, temporal flickering), measured via CLIP-I, optical flow, FID, and LPIPS; and (2) Video-Condition Alignment (object class, human action, spatial relationships, appearance style, physics, commonsense), assessed using CLIP text-image similarity, action recognition, grounded-SAM, and VQA models. The benchmark combines 50K+ human annotations with automated scoring for reliable evaluation. According to its standard procedure, for each prompt, sample 5 videos, for the Temporal Flickering dimension, sample 25 videos to ensure sufficient coverage after applying the static filter.

\paragraph{ChronoMagic-Bench.}ChronoMagic-Bench\cite{yuan2024chronomagicbenchbenchmarkmetamorphicevaluation} is a novel benchmark for evaluating text-to-time-lapse video generation models, assessing their ability to produce videos with significant metamorphic amplitude and temporal coherence. The dataset contains 1,649 prompt-video pairs categorized into four types (biological, human creation, meteorological, physical) and 75 subcategories. It introduces MTScore to measure metamorphic amplitude (combining video retrieval and GPT-4o scoring) and CHScore for temporal coherence (tracking point consistency and frame variations), while integrating existing UMT-FVD (visual quality) and UMTScore (text relevance) into a comprehensive four-dimensional evaluation framework.According to its standard procedure, for each prompt, sample 3 videos.
\subsection{Limitation}
\label{app:limitation}
A major limitation of our proposed method is its reliance on dataset construction and model training, which introduces additional training overhead compared to training-free methods. For instance, training our model on four NVIDIA A100 80G GPUs requires approximately one day.

\subsection{Visualization}
To provide a more comprehensive comparison, we present additional qualitative results of FastInit. These results clearly demonstrate that our method achieves remarkable effects in maintaining subject consistency and temporal continuity, thereby highlighting its superiority and practical value.

\begin{figure}
    \centering
    \includegraphics[width=1\linewidth]{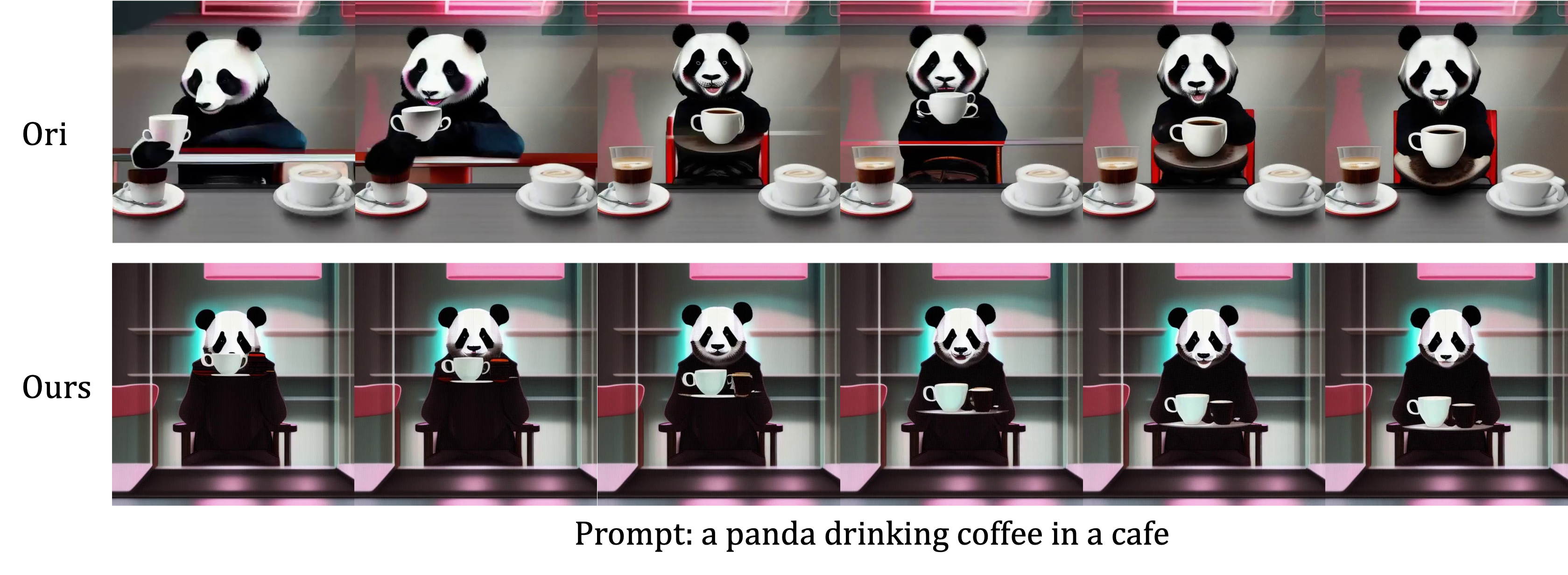}
\end{figure}
\begin{figure}
    \centering
    \includegraphics[width=1\linewidth]{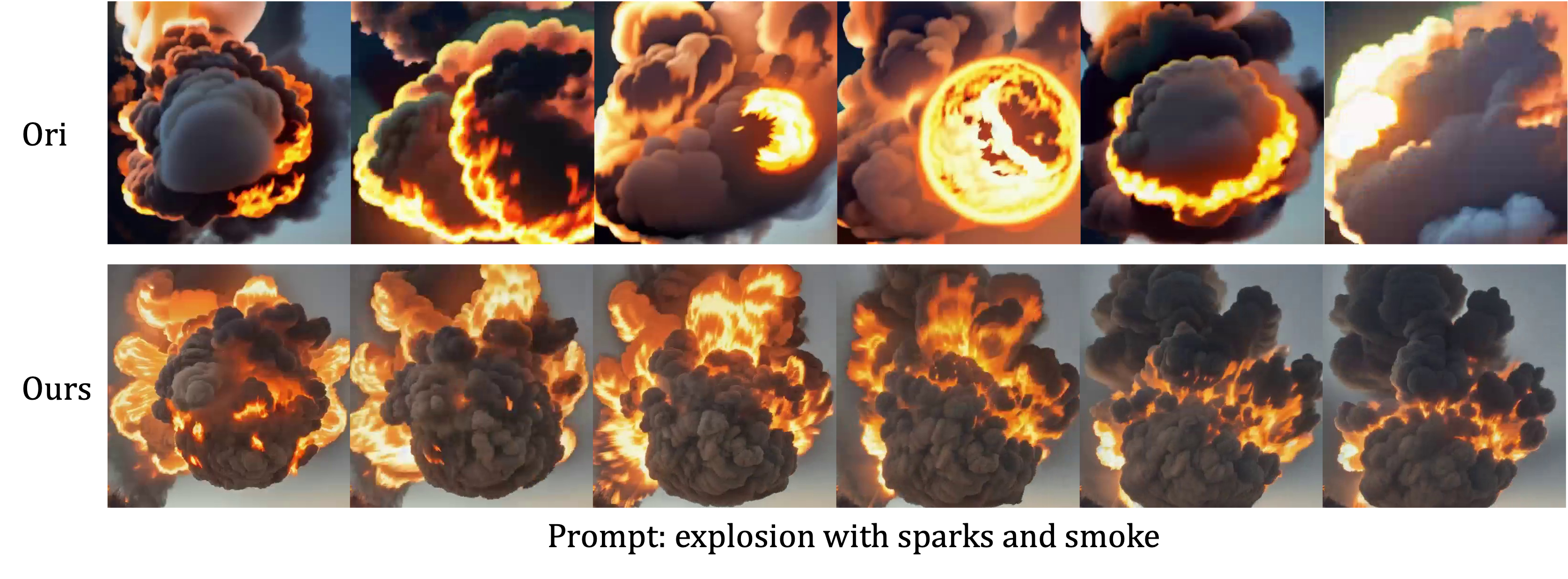}
\end{figure}
\begin{figure}
    \centering
    \includegraphics[width=1\linewidth]{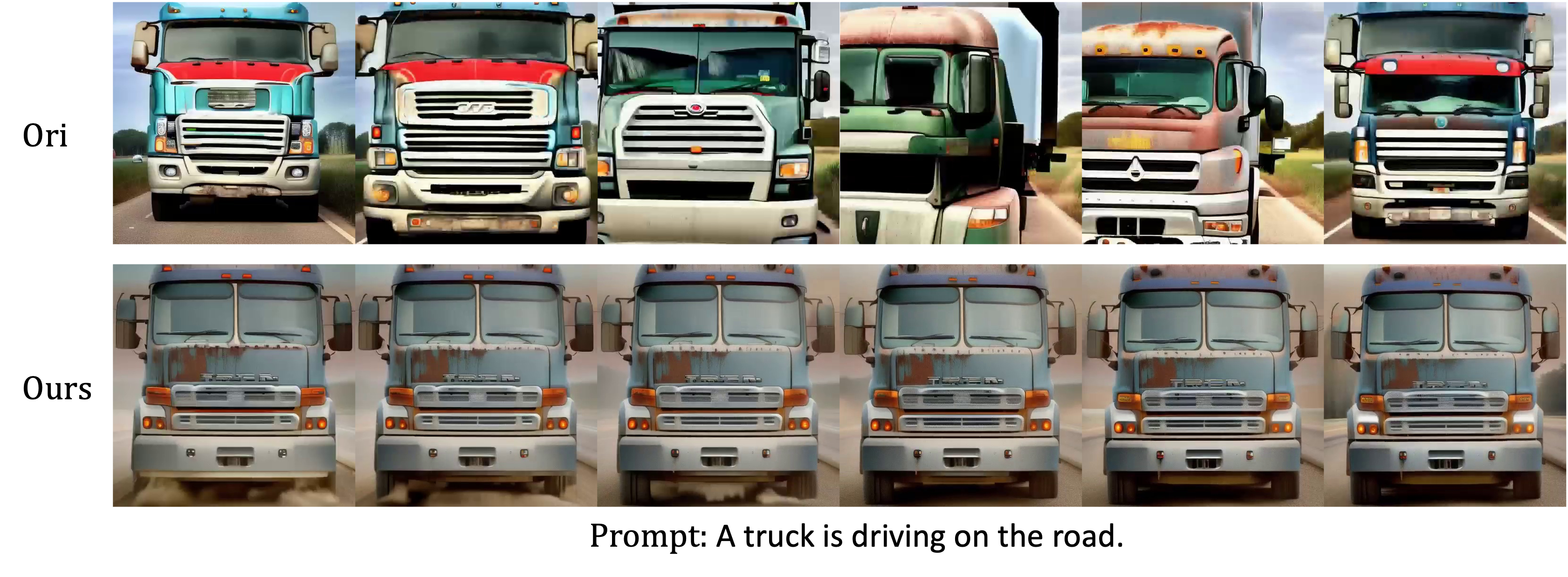}
\end{figure}
\begin{figure}
    \centering
    \includegraphics[width=1\linewidth]{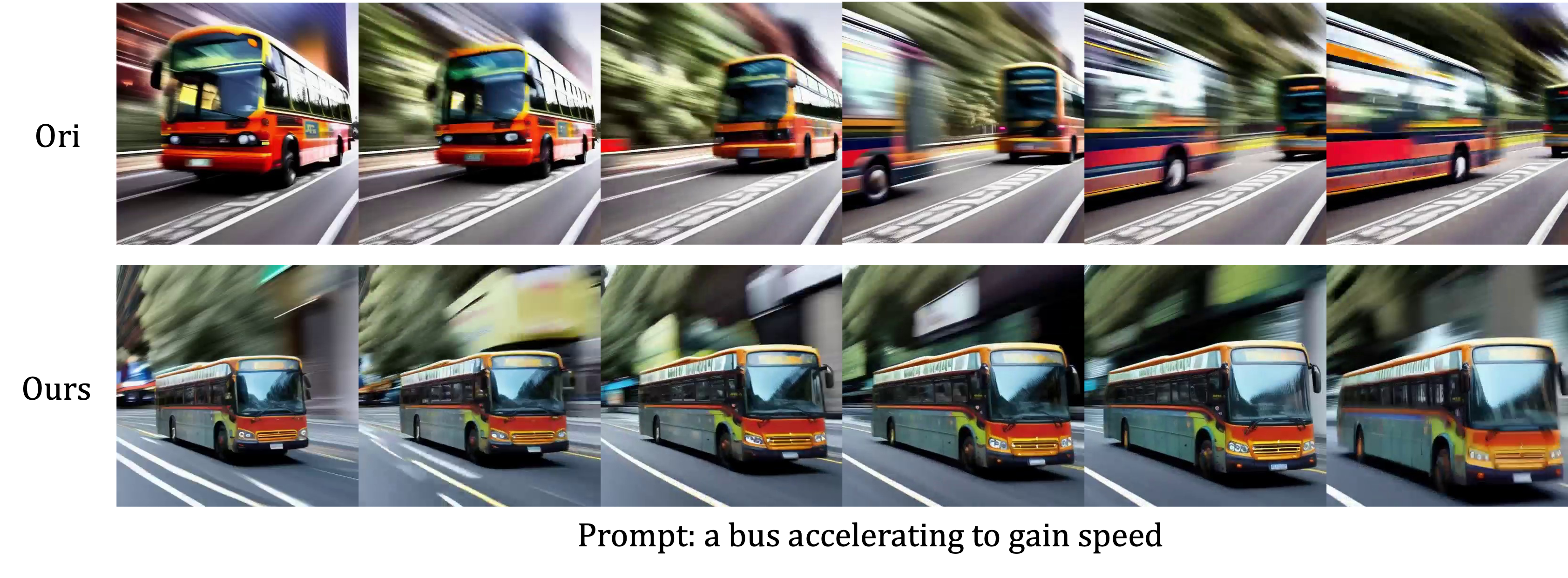}
\end{figure}
\begin{figure}
    \centering
    \includegraphics[width=1\linewidth]{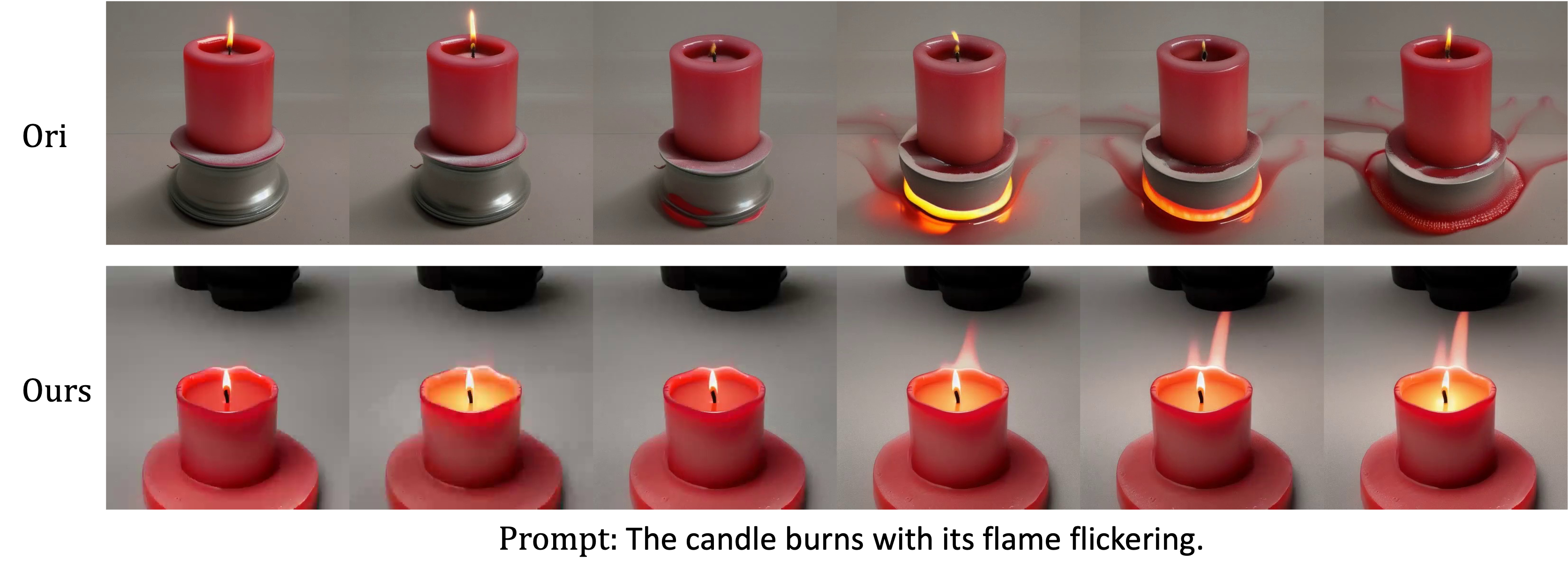}
\end{figure}
\begin{figure}
    \centering
    \includegraphics[width=1\linewidth]{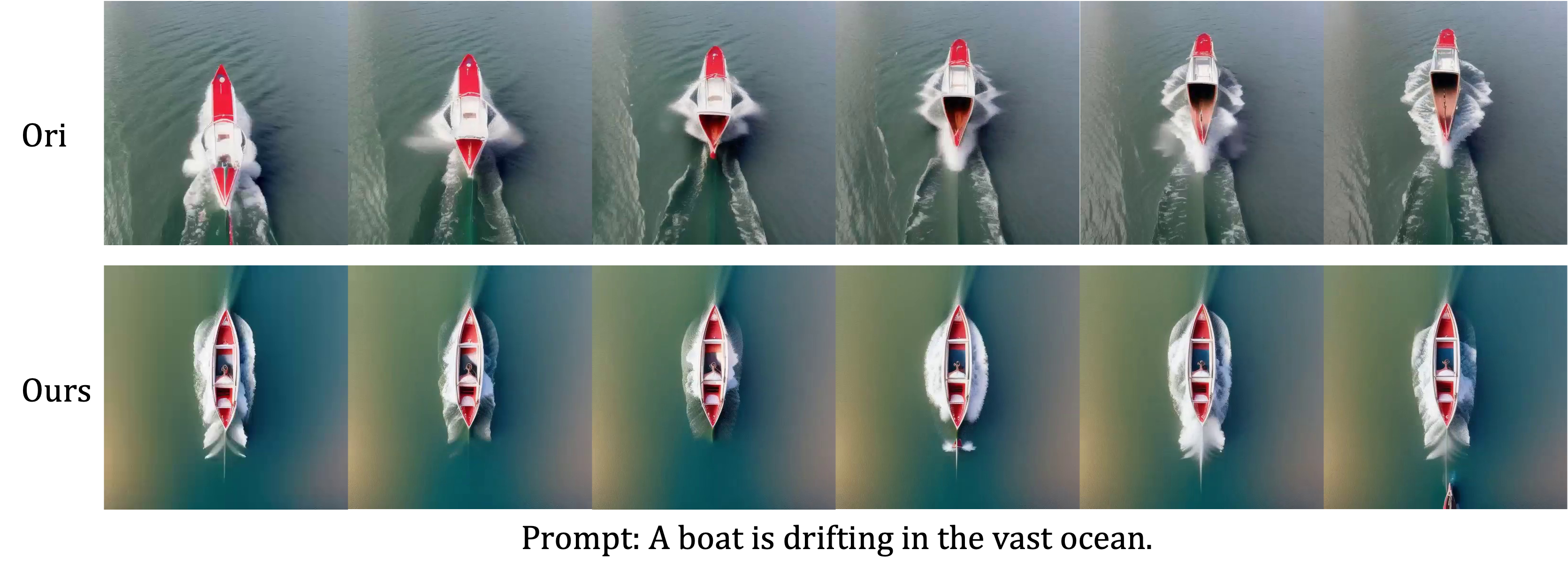}
\end{figure}
\begin{figure}
    \centering
    \includegraphics[width=1\linewidth]{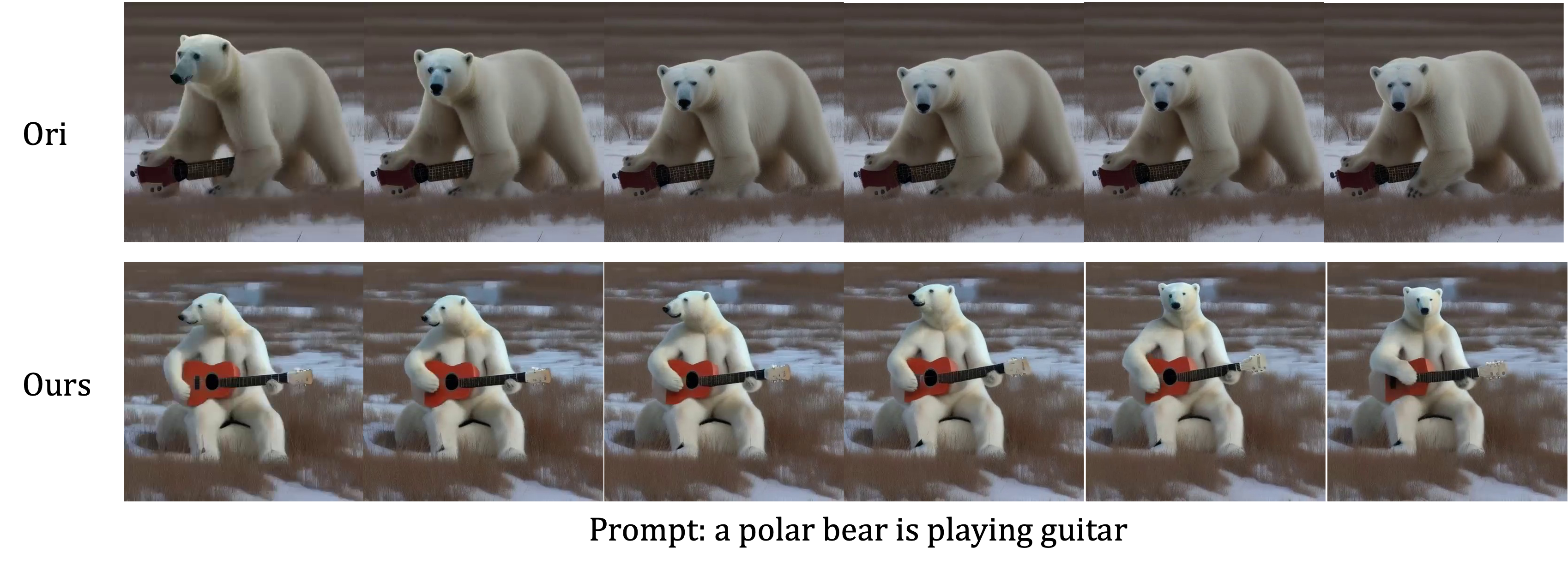}
\end{figure}

\begin{figure}
    \centering
    \includegraphics[width=1\linewidth]{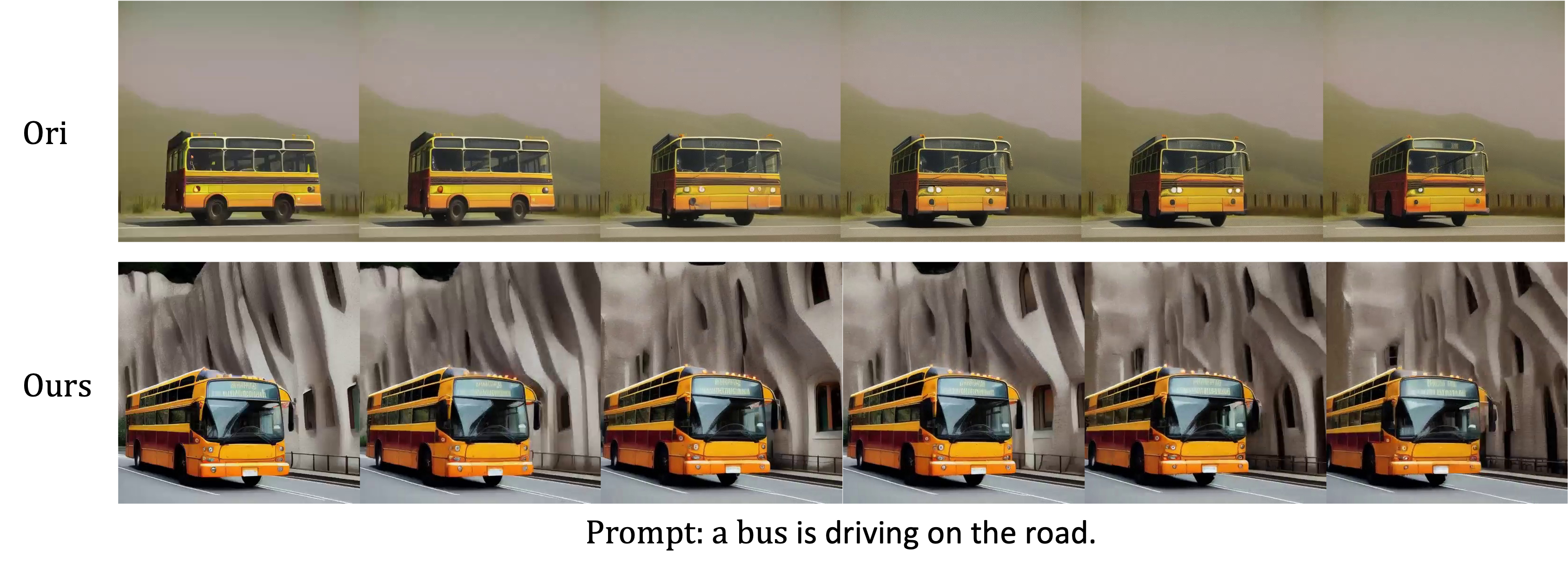}
\end{figure}

\end{document}